# AdaGScale: Viewpoint-Adaptive Gaussian Scaling in 3D Gaussian Splatting to Reduce Gaussian-Tile Pairs

Joongho Jo, Hyerin Lim, Hanjun Choi, and Jongsun Park
Korea University
Seoul, Republic of Korea
{jojoss1004, unk963, gkswns757, jongsun}@korea.ac.kr

## ABSTRACT

Reducing the number of Gaussian-tile pairs is one of the most promising approaches to improve 3D Gaussian Splatting (3D-GS) rendering speed on GPUs. However, the importance difference existing among Gaussian-tile pairs has never been considered in the previous works. In this paper, we propose AdaGScale, a novel viewpoint-adaptive Gaussian scaling technique for reducing the number of Gaussian-tile pairs. AdaGScale is based on the observation that the peripheral tiles located far from Gaussian center contribute negligibly to pixel color accumulation. This suggests an opportunity for reducing the number of Gaussian-tile pairs based on color contribution. AdaGScale efficiently estimates the color contribution in the peripheral region of each Gaussian during a preprocessing stage and adaptively scales its size based on the peripheral score. As a result, Gaussians with lower importance intersect with fewer tiles during the intersection test, which improves rendering speed while maintaining image quality. The adjusted size is used only for tile intersection test, and the original size is retained during color accumulation to preserve visual fidelity. Experimental results show that AdaGScale achieves a geometric mean speedup of 13.8× over original 3D-GS on a GPU, with only about 0.5 dB degradation in PSNR on city-scale scenes. The source code for AdaGScale is publicly available at https://github.com/askmgk/AdaGScale.

## 1 INTRODUCTION

3D Gaussian Splatting (3D-GS) [1]-[6] has recently emerged as a promising alternative to Neural Radiance Fields (NeRF) [7]-[10] for novel view synthesis. While NeRF represents scenes implicitly using neural networks and synthesizes images through neural inference, 3D-GS explicitly models scenes with millions of learnable 3D Gaussian primitives and synthesizes images through tile-based rendering. This explicit representation eliminates NeRF's complex sampling process and enables highly parallel rendering, resulting in substantially faster performance. Despite these advantages, 3D-GS still falls short of achieving the real-time rendering performance required for 3D computer vision applications such as augmented and virtual reality (AR/VR). For instance, even on Nvidia's server-grade A6000 GPU, the original 3D-GS implementation [1] achieves only around 10-15 frames per second (FPS) when rendering 4608×3456-resolution images. This is far below the real-time target of 90-120 FPS demanded by Apple Vision Pro, which features binocular displays with a combined resolution of 2×(3660×3200).

Recently, the methods for reducing the number of Gaussian-tile pairs [11]–[13] have attracted attention as GPU-friendly approaches that require no retraining, can be easily integrated with other optimization techniques. These methods primarily focus on losslessly eliminating unnecessary Gaussian-tile pairs by accurately computing the projected area of each Gaussian, thereby removing the pairs with exactly zero contribution to the final image. However, the variation in importance among pairs with non-zero contributions has not been considered.

Our analysis reveals that Gaussian-tile pairs generated during 3D-GS rendering vary significantly in their importance. Specifically, the tiles intersecting with the peripheral regions of each Gaussian contribute negligibly to pixel color accumulation. This observation suggests that selectively removing peripheral Gaussian-tile pairs based on their importance can significantly accelerate rendering with negligible image quality degradation. In this paper, we propose AdaGScale, a novel viewpoint-adaptive Gaussian scaling technique that leverages this observation to reduce the number of Gaussian-tile pairs. First, we introduce a peripheral score, defined as the sum of color contributions in the peripheral region of a Gaussian. Then, we present an approximate method to estimate this score during the preprocessing stage. During preprocessing, AdaGScale dynamically adjusts the size of each Gaussian in runtime based on peripheral score. This adjusted size is used only for the Gaussian-tile pair generation step, while the original size is retained for color accumulation in the rasterization stage. As a result, AdaGScale aggressively reduces the sizes of Gaussians with low peripheral scores, thereby reducing the number of Gaussian-tile pairs with them. This leads to substantial improvements in rendering speed while maintaining minimal degradation in image quality. Our main contributions are summarized as follows:

- We propose a method to approximate the peripheral score for Gaussian during the preprocessing stage.
- We present AdaGScale, a novel viewpoint-adaptive Gaussian scaling technique to eliminate low importance Gaussian-tile pairs based on the peripheral score.
- Through extensive experiments, we demonstrate that AdaGScale can be seamlessly integrated into existing 3D-GS pipelines without retraining or fine-tuning, and achieves substantial speedup on GPUs.

## 2 BACKGROUND AND MOTIVATION

### 2.1 Preliminaries of 3D Gaussian Splatting

3D Gaussian Splatting (3D-GS) [1] is an emerging technique in the field of novel view synthesis that explicitly models a scene using millions of learnable Gaussians. Fig. 1 illustrates the overall rendering pipeline of 3D-GS. The rendering pipeline consists of four main stages: preprocessing, Gaussian-tile pair generation, sorting, and rasterization.

In the preprocessing stage, each Gaussian is processed in parallel by a single GPU thread. As shown on the left side of Fig. 1, the surviving 3D Gaussians after culling step are projected onto the image plane. During this process, several features are computed, including projected 2D mean, projected 2D covariance, color, and depth. In the Gaussian-tile pair generation stage, the intersection of each 2D Gaussian is tested. For every intersecting tile, a Gaussian-tile pair is generated, consisting of tile ID, depth, and Gaussian ID. As shown in the example of Fig. 1, if a Gaussian intersects with five tiles, five corresponding Gaussian-tile pairs are generated.

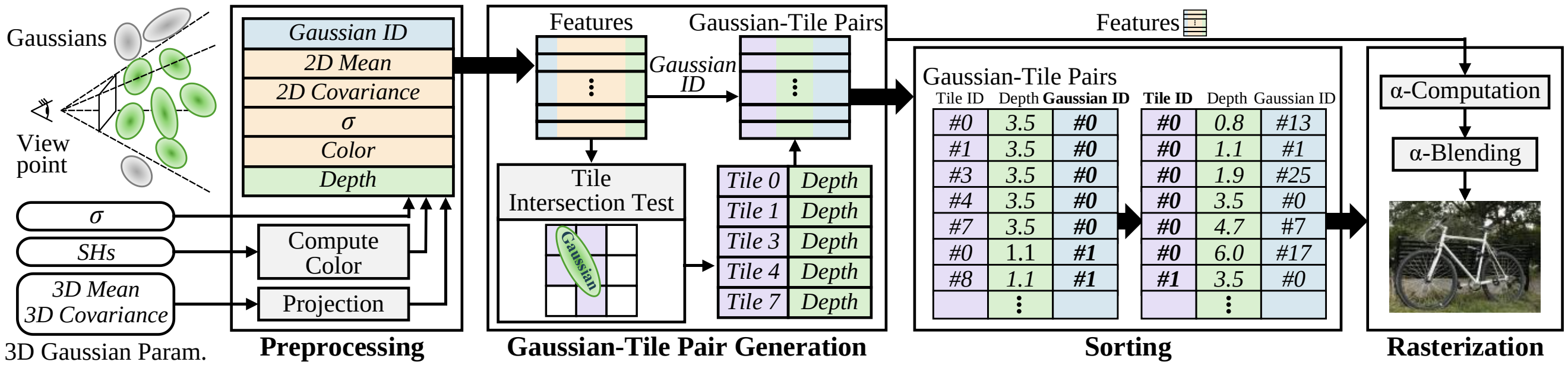

**Fig. 1: The overall rendering pipeline of 3D-GS.**

In the sorting stage, the generated Gaussian-tile pairs are sorted by depth to enable front-to-back rasterization within each tile. For efficient parallel sorting on GPUs, RadixSort [17] in NVIDIA CUB library is used.

In the rasterization stage, pixel colors are computed for each tile by processing its Gaussians in depth order with respect to current viewpoint. As shown in the right side of Fig. 1, the rasterization stage consists of two major steps. In $\alpha$-computation step, projected 2D mean ($G_{xy}$), projected 2D covariance ($G_{cov}$), and opacity ($\sigma$) of each Gaussian are used to determine its contribution to the color of the current pixel. The $\alpha$-value of $i$-th Gaussian is calculated as follows:

$$\alpha^i(P^k) = \sigma^i * \exp\left(-\frac{1}{2}\left(P^k_{xy} - G^i_{xy}\right)^T {G^i_{cov}}^{-1}\left(P^k_{xy} - G^i_{xy}\right)\right), \quad (1)$$

where $P^k_{xy}$ means $k$-th pixel's coordinates. In the $\alpha$-blending step, pixel colors are accumulated from front to back using the computed $\alpha$-values, as follows:

$$PixelColor(P^k) = \sum_{i=1}^{N_{pk}} \alpha^i(P^k) \cdot T^i(P^k) \cdot c^i, \quad (\alpha^i(P^k) \geq \tau), \quad (2)$$

where $T^i(P^k) = \prod_{j=1}^{i-1}\left(1 - \alpha^j(P^k)\right)$. Here, $N_{p^k}$ means the number of Gaussian associated $P^k$, and $\tau$ is the predefined $\alpha$ threshold, which is set to 1/255 in [1]. During the accumulation operation in (2), an early-exit is triggered if $T^i(P^k)$ falls below a predefined threshold, which is set to $10^{-4}$ in [1].

## 2.2 Bottleneck Analysis

To identify performance bottlenecks in 3D-GS rendering on GPUs, we conduct a detailed profiling analysis of the runtime breakdown across different rendering stages. The profiling is performed on NVIDIA A6000 GPU using six widely adopted 3D rendering scenes: Train, Truck, Drjohnson, Playroom, Bicycle, and Counter. Detailed descriptions of these scenes can be found in Section 4.1. Fig. 2 illustrates the detailed runtime breakdown of four rendering stages. Among these stages, the Gaussian-tile pair generation, sorting, and rasterization processes account for approximately 88.1% of total runtime on average. Since their computational workloads are largely determined by the number of Gaussian-tile pairs, reducing the pair count is a key factor in accelerating 3D-GS rendering.

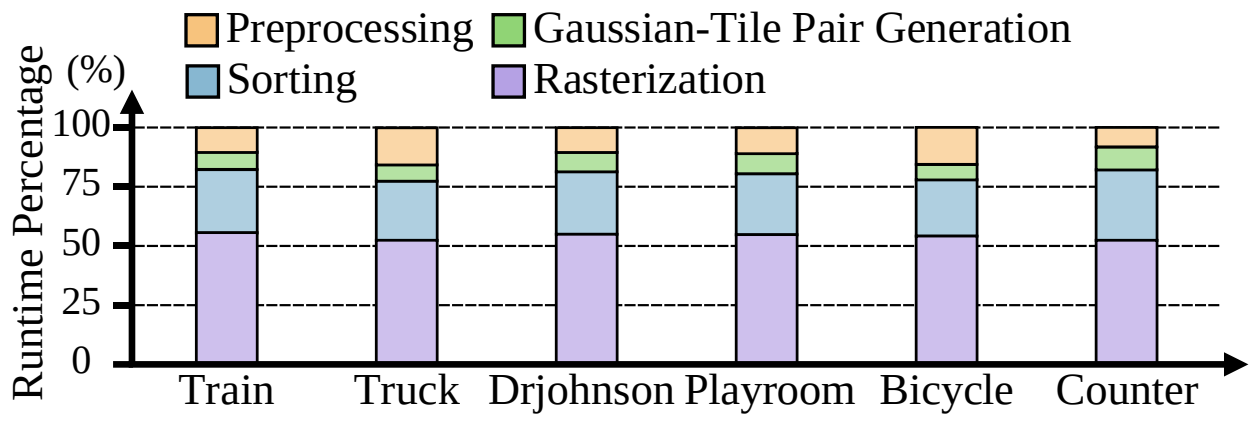

**Fig. 2: The normalized runtime breakdown of 3D-GS.**

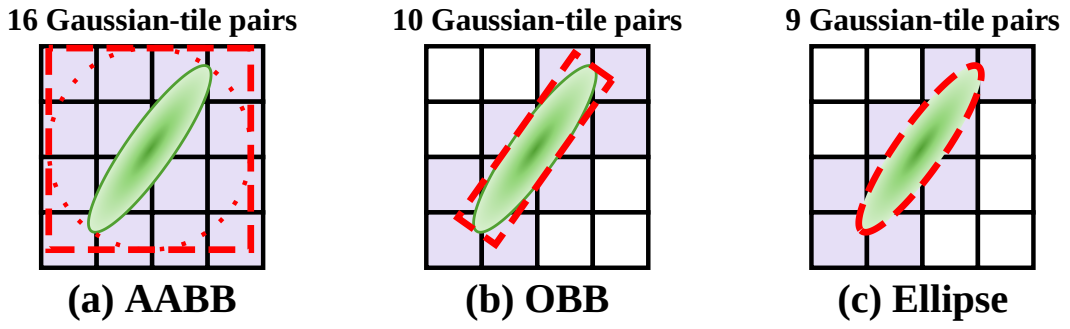

**Fig. 3: Comparison of tile intersection tests: (a) AABB, (b) OBB, and (c) Ellipse boundary.**

## 2.3 Related Works

Recent studies [11]-[13] have improved the rendering speed of 3D-GS by reducing Gaussian-tile pairs through more geometrically accurate identification of the tiles influenced by each Gaussian. Fig. 3 illustrates the approaches proposed in previous works for determining Gaussian-tile pairs. Fig. 3 (a) shows the axis-aligned bounding box (AABB) used in the original 3D-GS [1], Fig. 3 (b) depicts the oriented bounding box (OBB) employed in GSCore [11], and Fig. 3 (c) presents the ellipse-based boundary adopted in AdR-Gaussian [12] and FlashGS [13]. As shown in the figure, these geometry-based methods perform lossless reduction of Gaussian-tile pairs, which prevents them from considering the varying color contribution of individual pairs.

## 2.4 Color Contribution Analysis

The contribution of $i$-th Gaussian ($G^i$) to the color of $k$-th pixel ($P^k$) as adopted in previous works [4],[14]-[16] is given as the following:

$$Contribution(P^k, G^i) = \alpha^i(P^k) \cdot T^i(P^k). \quad (3)$$

Here, $Contribution(P^k, G^i)$ is derived from (2). As shown in (2), it represents the factor multiplied by the Gaussian color ($c^i$) in pixel color accumulation operation. Fig. 4 (a) shows the Peak Signal to Noise Ratio (PSNR) for four scenes when skipping the accumulation operations in (2) with ascending order of $Contribution(P^k, G^i)$. As shown in the graph, even when 80% of the operations are skipped, the PSNR decreases only by an average of 0.60 dB. This observation demonstrates that $Contribution(P^k, G^i)$ can serve as a meaningful metric for evaluating the importance of the pixel color accumulation operation. Additionally, we analyze the correlation between $Contribution(P^k, G^i)$ and the distance between the Gaussian center and the pixel. Fig. 4 (b) shows the variation of $Contribution(P^k, G^i)$ in relation to distance. As shown in the figure, $Contribution(P^k, G^i)$ decreases sharply as distance increases. This result suggests that the computations corresponding to the peripheral region of a Gaussian can be safely skipped based on their $Contribution(P^k, G^i)$, with negligible image quality degradation.

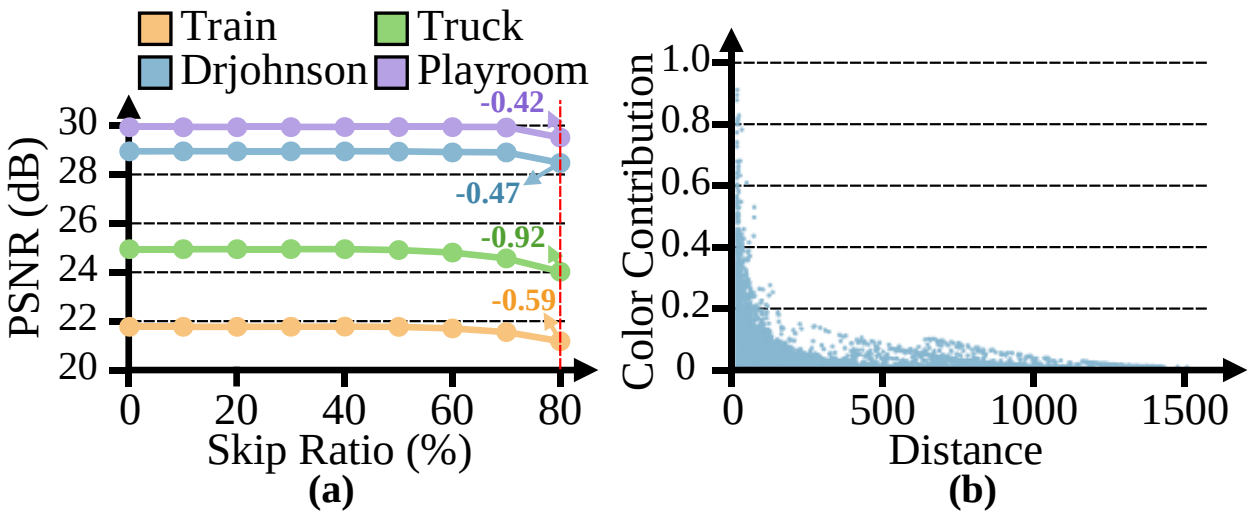

**Fig. 4: Analysis of the color contribution: (a) PSNR change when skipping low contribution operations, and (b) contribution decay with increasing Gaussian-pixel distance.**

# 3 PROPOSED WORK

## 3.1 Peripheral Score Definition

Based on the observation on $Contribution(P^k, G^i)$, we can define the peripheral score as the sum of color contributions from the pixels located in the peripheral region of Gaussian $G^i$. In the original 3D-GS [1], the predefined α threshold is set to 1/255, below which contributions can be ignored. We denote this threshold as $\tau = 1/255$. The peripheral region corresponds to the pixels where $\alpha$ value falls between $\tau$ and some intermediate threshold $x$, representing the area near the Gaussian's boundary. The peripheral score can be defined as follows:

$$PS(G^i, x) = \sum_{\tau \le \alpha^i(P^k) < x} Contribution(P^k, G^i). \quad (4)$$

The peripheral score represents how much a Gaussian contributes to the pixels near its boundary. To control the trade-off between rendering speed and quality, we introduce $K$, a user-defined parameter representing the acceptable peripheral loss. Further details on $K$ are provided in subsection 3.4. $PS(G^i, x)$ is also employed to determine the maximum $\alpha$ threshold at which peripheral contributions can be ignored while maintaining image quality. Therefore, we define $Th^i$ as the largest α threshold $x$ that satisfies $PS(G^i, x) \le K$, meaning that the accumulated contribution in the peripheral region ($\tau \le \alpha < Th^i$) remains below the predefined $K$, as follows:

$$Th^i = max\{x | PS(G^i, x) \le K\}. \quad (5)$$

By setting $Th^i$ as the $\alpha$ threshold for Gaussian $i$, instead of using the fixed value of 1/255, we can effectively adjust its coverage area. If $Th^i$ is larger than 1/255, the $\alpha$ threshold for Gaussian $i$ increases from 1/255 to $Th^i$, which reduces its coverage area. The tile intersection test is then performed using this reduced coverage, decreasing the number of Gaussian-tile pairs that need to be evaluated during rendering. Conversely, if $Th^i$ is smaller than 1/255, Gaussian $i$ is considered important for rendering quality, and the default threshold of 1/255 is maintained as in the original 3D-GS.

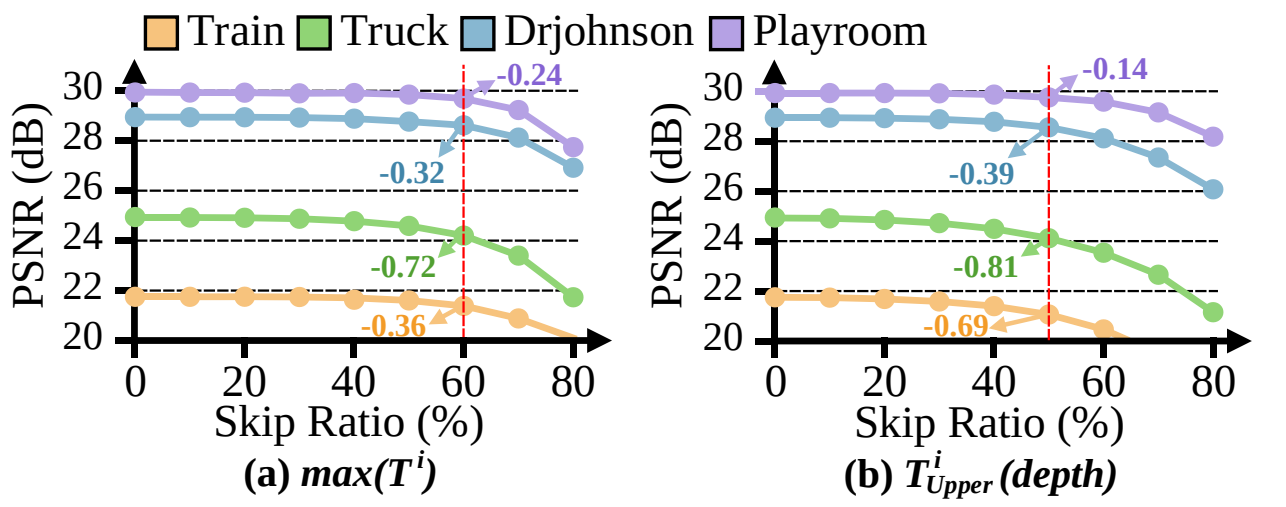

**Fig. 5: Ablation study on approximation stages. PSNR change when skipping low contribution operations based on (a) $max(T^i)$approximation and (b) $T^i_{Upper}(depth)$.**

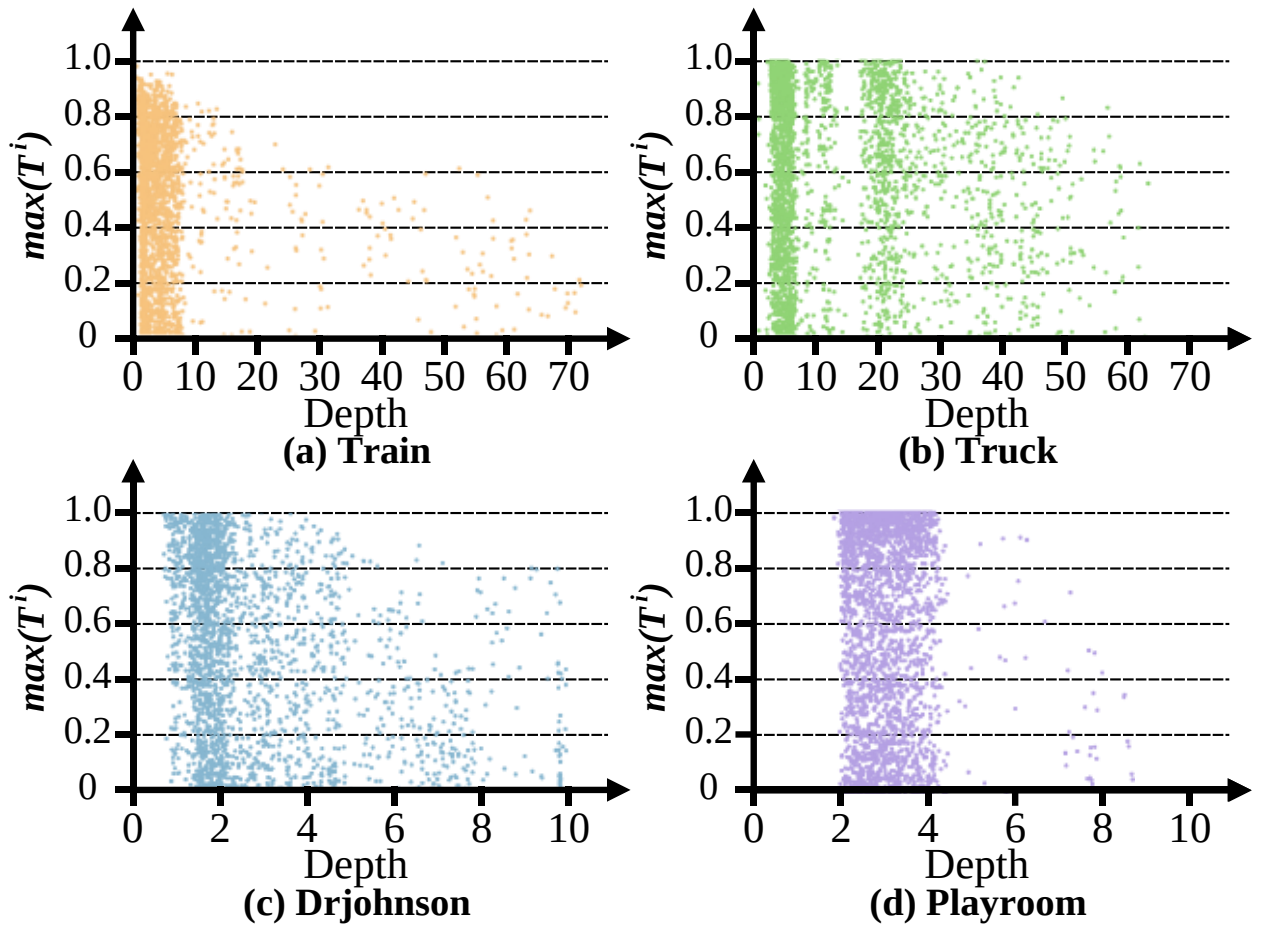

**Fig. 6: Distribution of $max(T^i)$ and depth across different scenes.**

**Challenge.** In practice, however, computing $PS(G^i, x)$ directly during the preprocessing stage is infeasible since it requires both $\alpha^i$ and $T^i$ values (writing $\alpha^i(P^k)$ and $T^i(P^k)$ simply as $\alpha^i$ and $T^i$ for clarity). One of our primary goals is to reduce the number of Gaussian-tile pairs, thereby minimizing expensive $\alpha^i$ computations. So, calculating $\alpha^i$ to calculate $PS(G^i, x)$ would contradict this goal. Furthermore, $T^i$ represents the cumulative transmittance of Gaussians preceding current one, which can only be accurately calculated after the sorting operation. Therefore, to utilize the peripheral score before the Gaussian-tile pair generation stage, it is essential to develop an efficient method to approximate $PS(G^i, x)$ during the preprocessing stage with minor computational overhead.

## 3.2 Efficient Approximation of Peripheral Score

To enable practical computation of the peripheral score during preprocessing, we develop a three-stage approximation strategy.

**Transmittance Approximation.** We first address the challenge of approximating $T^i$, which depends on the depth-based sorting order and thus cannot be computed accurately during preprocessing. Our key insight is that replacing the pixel-specific transmittance $T^i$ with a single constant value per Gaussian can incur only negligible loss in contribution accuracy. Specifically, we investigate whether $T^i$ can be approximated using the maximum value across all the pixels influenced by the Gaussian (i.e., $max(T^i)$). To validate this approximation, we conduct the same experimental setup as Fig. 4 (a), but using the approximated contribution $max(T^i) \cdot \alpha^i$ instead of the exact value $T^i \cdot \alpha^i$. Fig. 5 (a) demonstrates that even when 60% of the computations are skipped in the ascending order of approximated contribution, the average PSNR decreases by merely 0.41 dB. Compared to the exact contribution in Fig. 4 (a), this result indicates that $T^i$ can be effectively approximated as a single constant per Gaussian, as follows:

$$PS(G^i, x) \approx max(T^i(P^k)) \cdot \sum_{\tau \leq \alpha^i(P^k) < x} \alpha^i(P^k). \quad (6)$$

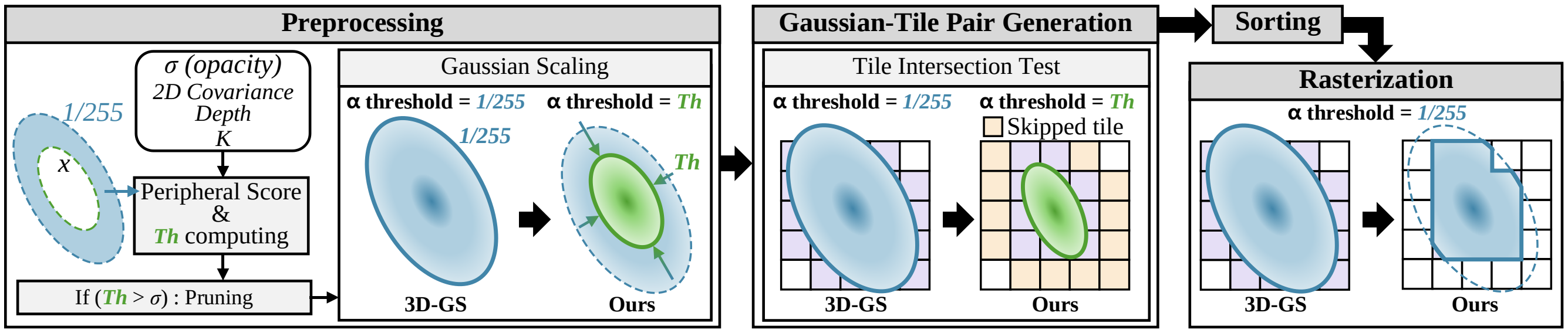

**Fig. 7: Overview of AdaGScale pipeline and its comparison with 3D-GS.**

**Depth-Based Upper Bound Estimation.** Having established that $max(T^i)$ can serve as a suitable approximation, we now address how to estimate this value during preprocessing. Among the features computed in preprocessing—such as projected 2D mean, projected 2D covariance, opacity, and depth—depth is the only parameter that directly influences the value of $T^i$. This is because $T^i$ is computed based on the depth-based sorting order. We analyze the correlation between depth and $max(T^i)$ across multiple scenes. Fig. 6 shows the distribution of depth and $max(T^i)$ for Train, Truck, Drjohnson, and Playroom scenes. As observed in the figures, $max(T^i)$ generally decreases as depth increases, which aligns with the intuition that Gaussians farther from the camera are more likely to be occluded. However, this relationship varies significantly across scenes, and small depth values do not always correspond to large $max(T^i)$ values. For instance, in the playroom scene (Fig. 6 (d)), most depth values lie within the range of 2 to 4, while $max(T^i)$ is spread across the entire range from 0 to 1. In such cases, depth alone is insufficient for accurately predicting $T^i$. Therefore, we employ a piecewise constant upper bound of $max(T^i)$ as a function of depth, denoted as $T^i_{Upper}(depth)$. Since this upper bound differs across scenes, it can be obtained offline using a calibration dataset randomly sampled training dataset. $T^i_{Upper}(depth)$ is described further in subsection 3.4. To assess the accuracy of this approximation, we repeat the evaluation from Fig. 4 (a) using $T^i_{Upper}(depth) \cdot \alpha^i$ as the approximated contribution instead of the exact value $T^i \cdot \alpha^i$. Fig. 5 (b) demonstrates that even when 50% of the computations are skipped in the ascending order of approximated contribution, the average PSNR decreases by about 0.51 dB. The approximated $PS(G^i, x)$ can be expressed as follows:

$$PS(G^i, x) \approx T^i_{Upper}(depth) \cdot \sum_{\tau \leq \alpha^i(P^k) < x} \alpha^i(P^k). \quad (7)$$

**Closed-Form Computation via Riemann Sum.** Finally, a closed-form expression is derived for the efficient computations of $\sum_{\tau \leq \alpha^i(P^k) < x} \alpha^i(P^k)$ in (7). The summation represents an accumulation over discrete pixel locations within a continuous 2D spatial domain. To enable efficient computation, we approximate this discrete sum using a Riemann sum formulation:

$$PS(G^i, x) \approx T^i_{Upper}(depth) \cdot \int_{\tau \leq \alpha^i(P^k) < x} \alpha^i(P^k) dP^k. \quad (8)$$

For a 2D Gaussian with covariance $G^i_{cov}$, the integral over the annular region, where $\tau \leq \alpha^i(P^k) < x$ can be evaluated in polar coordinates using standard Gaussian integration. This results in the following form:

$$PS(G^i, x) \approx T^i_{Upper}(depth) * 2\pi\sqrt{det(G^i_{cov})}(x - \tau). \quad (9)$$

Since the approximated peripheral score in (9) is a monotonically increasing function of $x$, it can be substituted into (5) to solve for $Th^i$, as follows:

$$Th^i = \frac{K}{T^i_{Upper}(depth) \cdot 2\pi\sqrt{det(G^i_{cov})}} + \tau, \quad (10)$$

which provides a closed-form expression for $Th^i$ that depends only on the parameters available during preprocessing: depth, projected 2D covariance ($G^i_{cov}$), and the predefined $K$. Crucially, both depth and $G^i_{cov}$ vary with a particular viewpoint, making $Th^i$ viewpoint-adaptive, and it is dynamically recomputed for each viewpoint during preprocessing with negligible overhead.

## 3.3 Overall Pipeline of AdaGScale

This subsection presents a viewpoint-adaptive dynamic scaling technique that leverages the peripheral score to efficiently reduce the number of Gaussian-tile pairs. Fig. 7 illustrates the overall pipeline of AdaGScale and its comparison with the original 3D-GS.

**Preprocessing Stage.** As illustrated in Fig. 7, the preprocessing stage computes $Th^i$ for each Gaussian based on $G^i_{cov}$, depth, and the predefined $K$ using (10). As shown in (10), since $K$, $T^i_{Upper}(depth)$, and $det(G^i_{cov})$ are all positive values, $Th^i$ is always greater than or equal to the original α threshold $\tau$ (= 1/255). We then set the $\alpha$ threshold of the $i$-th Gaussian to $Th^i$. An important point here is that if $Th^i$ is greater than the opacity $\sigma^i$ of the Gaussian, the Gaussian is considered to have negligible influence on the current viewpoint and is excluded from subsequent computations. Otherwise, the Gaussian proceeds to the Gaussian-tile pair generation stage with the adjusted threshold $Th^i$.

**Gaussian-Tile Pair Generation Stage.** During the Gaussian-tile pair generation stage, the adjusted $\alpha$ threshold $Th^i$ is applied to the tile intersection test. As the threshold increases beyond $\tau$, the effective coverage area of each Gaussian is reduced, resulting in fewer intersecting tiles as shown in the middle panel of Fig. 7, where some tiles are skipped. Consequently, the reduced number of

TABLE I. RESOLUTION OF DATASETS

| Dataset | Scene | Resolution |
|---|---|---|
| T&T | Train, Truck | 980×545 |
| DB | Drjohnson, Playroom | 1332×876, 1264×832 |
| Mip360 | Bicycle, Bonsai, Counter, Flowers, Garden, Kitchen, Room, Stump, Treehill | 1245×825 to 1559×1039 |
| M19 | Building, Rubble | 4608×3456 |
| US3D | Residence, Sci-Art | 5472×3648, 4864×3648 |

Gaussian-tile pairs significantly lowers the computational complexity of both the sorting and rasterization stages.

**Rasterization Stage.** One of the key points of AdaGScale is the proposed dual-size strategy. As shown in Fig. 7, while the adjusted threshold $Th^i$ is used during pair generation to reduce the number of pairs, the rasterization stage always uses the original $\alpha$ threshold $\tau$ (= 1/255) when computing pixel colors. The purpose of dynamically adjusting the threshold is not to suppress the true influence of a Gaussian, but rather to identify and eliminate low-importance Gaussian-tile pairs during the intersection test. Therefore, we keep the original $\alpha$ threshold during color accumulation so that any Gaussian that remains associated with a tile, contributes exactly with its original, unaltered $\alpha$ value. This dual-size approach enables aggressive pair reduction while preserving accurate color computation for the remaining pairs, thereby achieving substantial speedup with minimal quality degradation.

## 3.4 Peripheral Loss K and Upper Bound of T

Both $K$ and $T^i_{Upper}$ are determined offline using 16 viewpoints sampled from the training dataset. For $K$, binary search is used with a target PSNR degradation threshold, completing in approximately 10-30 seconds per scene. This efficiency arises from the proportional relationship: larger $K$ eliminates more Gaussian-tile pairs, proportionally increasing PSNR loss. For $T^i_{Upper}$, we use a piecewise constant upper bound by dividing the depth range [0, 100] into 20 uniform intervals, assigning each the maximum $T^i$ from the calibration dataset. Implementation on GPU uses a Look-Up Table (LUT) for instant $T^i_{Upper}$ retrieval given depth. Determining $T^i_{Upper}$ also takes approximately 10-30 seconds per scene.

# 4 EXPERIMENTAL RESULTS

## 4.1 Experimental Setup

**Implementation.** AdaGScale is implemented using CUDA programming by modifying the author-released codes from FlashGS [13] and Scaffold-GS [19]. All experiments are conducted on an NVIDIA A6000 GPU. The code is compiled using GCC 11.4.0 and NVCC from CUDA 11.6. **Datasets.** We evaluate our method on diverse real-world indoor and outdoor scenes from five benchmark datasets: Tanks&Temples (T&T) [20], DeepBlending (DB) [21], Mip-NeRF360 (Mip360) [22], Mill-19 (M19) [23], and UrbanScene3D (US3D) [24]. For T&T, DB, and Mip360, every 8th image is used for testing. For M19, every 64th image is used for testing, and for US3D, every 128th image is used for testing. Table I shows the resolution and scene details of the evaluated datasets. **Gaussian Models.** We evaluate the performance of AdaGScale using five Gaussian splatting models: original 3D-GS [1], GaussianSpa [5], CityGaussian [18], and Scaffold-GS [19]. GaussianSpa is a pruned model that represents 3D scenes with approximately 8× fewer Gaussians on average compared to original 3D-GS. CityGaussian is trained on high-resolution city-scale scenes.

TABLE II. RENDERING QUALITY OF BASELINE

**Original 3D-GS (dB)**

| Scene | Train | Truck | Drjohn. | Playroom | Bicycle | Bonsai | Counters |
|---|---|---|---|---|---|---|---|
| PSNR | 21.79 | 24.97 | 28.92 | 29.95 | 25.14 | 31.92 | 28.84 |
| Scene | Flowers | Garden | Kitchen | Room | Stump | Treehill | |
| PSNR | 21.46 | 27.19 | 30.67 | 31.28 | 26.58 | 22.34 | |

**GaussianSpa (dB)**

| Scene | Train | Truck | Drjohn. | Playroom | Bicycle | Bonsai | Counters |
|---|---|---|---|---|---|---|---|
| PSNR | 21.41 | 25.36 | 29.43 | 30.50 | 25.31 | 31.23 | 28.67 |
| Scene | Flowers | Garden | Kitchen | Room | Stump | Treehill | |
| PSNR | 21.63 | 27.07 | 30.93 | 31.29 | 27.10 | 23.00 | |

**Scaffold-GS (dB)**

| Scene | Train | Truck | Drjohn. | Playroom |
|---|---|---|---|---|
| PSNR | 22.21 | 25.80 | 29.72 | 30.81 |

**CityGaussian (dB)**

| Scene | Building | Rubble | Residence | Sci-Art |
|---|---|---|---|---|
| PSNR | 19.48 | 22.01 | 20.17 | 20.92 |

TABLE III. AVERAGE ACTUAL PSNR DROP FOR EACH GAUSSIAN MODEL

| Target PSNR drop | 0.1 | 0.2 | 0.3 | 0.4 | 0.5 |
|---|---|---|---|---|---|
| Original 3D-GS | 0.053 | 0.104 | 0.157 | 0.210 | 0.264 |
| GaussianSpa | 0.083 | 0.165 | 0.244 | 0.321 | 0.398 |
| Scaffold-GS | 0.094 | 0.182 | 0.268 | 0.351 | 0.434 |
| CityGaussian | 0.082 | 0.165 | 0.248 | 0.331 | 0.415 |

Scaffold-GS is a novel 3D-GS rendering pipeline that generates and renders Gaussians in real-time using a multi-layer perceptron (MLP). We use the pre-trained Gaussian models publicly released by the authors for original 3D-GS and CityGaussian. For compressed models where pre-trained models are not available, such as GaussianSpa, and Scaffold-GS, we train the models for 30K iterations using the author-released codes. **Evaluation Metrics.** Rendering quality is measured using PSNR, and rendering speed is measured in FPS.

## 4.2 Analysis of PSNR Drop

Although AdaGScale determines the PSNR drop through a predefined $K$, as explained in subsection 3.4, $K$ is determined using the training dataset, so the actual PSNR drop on the test dataset may differ from the target PSNR drop. Therefore, we analyze the actual PSNR drop according to the target PSNR drop. Table II shows the baseline PSNR of the Gaussian models used in the evaluation. Table III presents the average actual PSNR drop for each Gaussian model according to the target PSNR drop. As can be seen from the results, the actual PSNR drop does not exceed the target value in all cases.

## 4.3 Speedup Evaluation on Various Models

As shown in the previous subsection, when $K$ is determined using the target PSNR drop on the calibration dataset, the actual PSNR drop on the test dataset exhibits some deviation from the target value. This deviation can vary across different scenes, potentially introducing confounding factors when comparing speedups. For example, if the same target PSNR drop results in a 0.1 dB drop for scene A but a 0.4 dB drop for scene B, comparing the speedups between these scenes would be unfair as they incur different PSNR costs. Therefore, to ensure fair comparison of pure speedup gains at exactly identical PSNR drop levels (e.g., -0.5 dB) on the test dataset, we fine-tune $K$ using the test dataset for this evaluation. This guarantees that all compared methods operate at the same quality level, eliminating the influence of $K$ calibration errors.

Fig. 8 shows the speedup comparison across different Gaussian splatting models: (a) original 3D-GS, (b) the pruned model

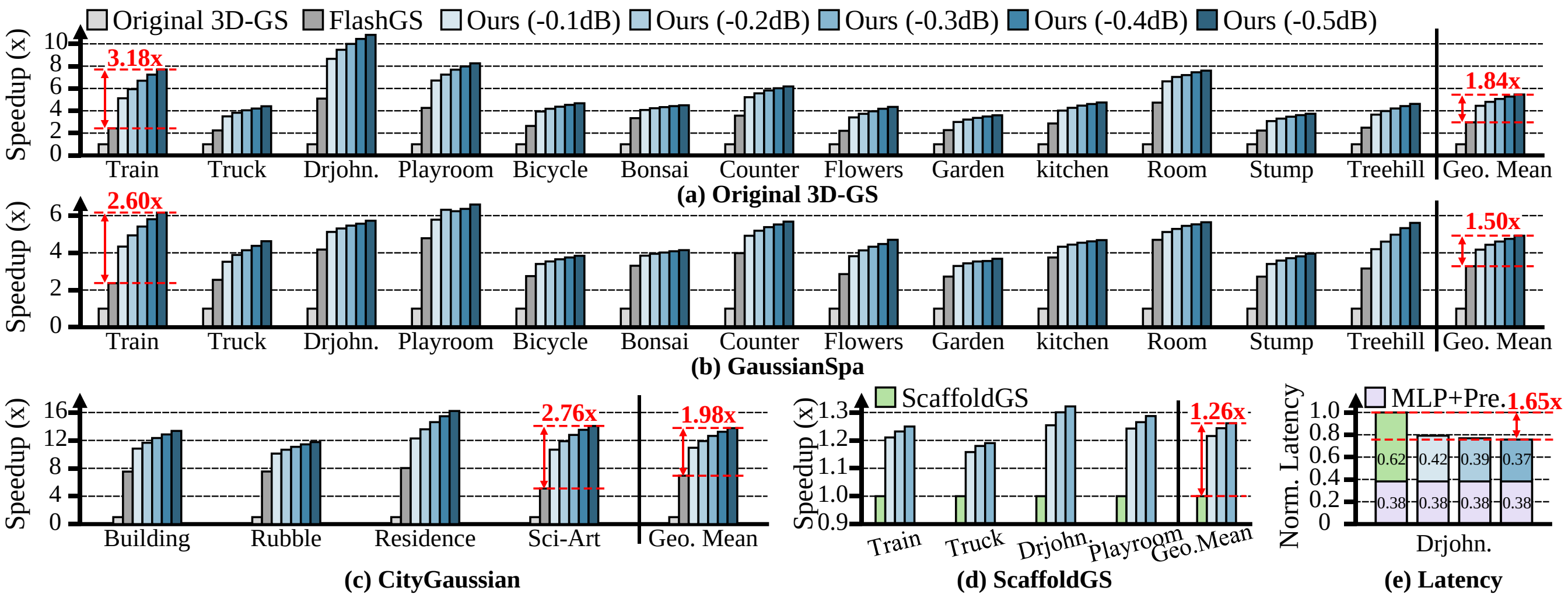

**Fig. 8: Speedup comparison across different Gaussian splatting models: (a) original 3D-GS, (b) GaussianSpa, (c) CityGaussian, and (d) Scaffold-GS. (e) shows the latency breakdown with and without AdaGScale on Scaffold-GS.**

GaussianSpa, (c) the city-scale model CityGaussian, and (d) the MLP-based model Scaffold-GS. We compare three rendering pipelines: original 3D-GS, FlashGS, and AdaGScale (Ours), where the PSNR drop values in parentheses (e.g., -0.1 dB) indicate the actual PSNR drop measured on the test dataset. Fig. 8 (e) shows the latency breakdown with and without AdaGScale applied to Scaffold-GS on the Drjohnson scene.

**Original 3D-GS and GaussianSpa.** As shown in Fig. 8 (a) and (b), at 0.5 dB PSNR drop, AdaGScale achieves geometric mean speedups of 5.46× and 4.92× over the original 3D-GS pipeline for original 3D-GS and GaussianSpa models, respectively, and 1.84× and 1.50× over FlashGS. These results demonstrate that AdaGScale can achieve substantial speedups with negligible PSNR degradation, even on Gaussian models already compressed through pruning.

**CityGaussian.** Fig. 8 (c) shows the speedup comparison using CityGaussian. At 0.5 dB PSNR drop, AdaGScale achieves a geometric mean speedup of 13.8× over original 3D-GS and 1.98× over FlashGS across four scenes. These results demonstrate that AdaGScale can achieve substantial speedups not only on small-scale scenes but also on city-scale high-resolution scenes.

**Scaffold-GS.** Fig. 8 (d) shows the comparison results using Scaffold-GS, a novel 3D-GS rendering pipeline that generates and renders Gaussians in real-time using a multi-layer perceptron (MLP). AdaGScale achieves a geometric mean speedup of 1.26× over Scaffold-GS at 0.3 dB PSNR drop across four scenes. Fig. 8 (e) provides further insight into the speedup breakdown. The MLP operations for Gaussian generation and preprocessing operations for voxel-level culling, which are absent in the original 3D-GS rendering, account for 38% of the total latency. The actual rendering stage accounts for 62%, and when AdaGScale is applied, considering only the rendering portion, it achieves a 1.65× speedup at 0.3 dB PSNR drop. This demonstrates that AdaGScale can effectively achieve speedups even on emerging MLP-based 3D-GS rendering pipelines.

### 4.4 Analysis of Gaussian-Tile Pairs

Following the same approach as the speedup evaluation, we fine-tuned $K$ using the test dataset to align the actual PSNR drop with the target value. Table IV presents the average reduction ratio of Gaussian-tile pairs during rendering compared to FlashGS for each Gaussian model across different target PSNR drops. As shown in the table, AdaGScale achieves up to 43.07% reduction in Gaussian-tile pairs for the Original 3D-GS model. This substantial reduction in Gaussian-tile pairs directly translates to the observed speedups in Fig. 8, as fewer pairs result in reduced computational costs in the sorting and rasterization stages.

TABLE IV. GAUSSIAN-TILE REDUCTION COMPARED TO FLASHGS

| % | 0.1 | 0.2 | 0.3 | 0.4 | 0.5 |
|---|---|---|---|---|---|
| **Original 3D-GS** | 31.03 | 35.77 | 38.88 | 41.19 | 43.07 |
| **GaussianSpa** | 20.20 | 24.15 | 26.91 | 29.09 | 30.93 |
| **CityGaussian** | 19.88 | 25.33 | 28.88 | 31.54 | 33.64 |

## 5 CONCLUSION

In this paper, we investigate the varying importance of Gaussian-tile pairs in 3D Gaussian Splatting and identify that peripheral pairs contribute negligibly to image quality. To address this, we present AdaGScale, a viewpoint-adaptive Gaussian scaling technique that accelerates rendering by selectively reducing low-importance pairs based on our peripheral score metric. Our key contributions include: (1) an efficient approximation method for estimating peripheral scores during preprocessing, and (2) a dual-size strategy that reduces Gaussian-tile pairs while preserving rendering accuracy. AdaGScale requires no retraining and seamlessly integrates with existing 3D-GS pipelines. Experimental results across diverse Gaussian models and scene types demonstrate that AdaGScale achieves substantial speedups—5.46× geometric mean on original 3D-GS, up to 13.8× on city-scale scenes.

## ACKNOWLEDGMENT

This work was supported in part by the National Research Foundation of Korea(NRF) grant funded by the Korea government(MSIT) (RS-2024-00345481) and (No.RS-2024-00405495, Plug&Play(P&P) Chiplet Integration research center), and in part by Ministry of Trade, Industry and Energy(MOTIE)and Korea Institute for Advancement of Technology(KIAT) through the "International Cooperative R&D program"(Task No. P0028486).